\documentclass[letterpaper, 10 pt, conference]{ieeeconf}  
\usepackage{graphicx}
\usepackage{amsmath}
\usepackage{adjustbox}
\usepackage{xcolor}
\usepackage{stmaryrd}
\usepackage{hyperref}
\let\labelindent\relax
\usepackage{enumitem}
\usepackage{wrapfig}
\usepackage{mathtools}
\usepackage[linesnumbered,algoruled,boxed,lined]{algorithm2e}
\usepackage{algpseudocode}
\usepackage[utf8]{inputenc} 
\usepackage[T1]{fontenc}    
\usepackage{hyperref}       
\usepackage{url}            
\usepackage{booktabs}       
\usepackage{amsfonts}       
\usepackage{nicefrac}       
\usepackage{microtype}      
\usepackage{xspace}
\usepackage{mathtools, nccmath}
\newcommand{\Z}{\mathbb Z}
\usepackage{multirow}
\usepackage{cite}

\usepackage{float}
\usepackage{caption}
\usepackage{ctable}
\graphicspath{{fig/}}
\renewcommand{\baselinestretch}{0.93}

\newtheorem{definition}{Definition}[section]

\AtBeginDocument{%
  \providecommand\BibTeX{{%
    \normalfont B\kern-0.5em{\scshape i\kern-0.25em b}\kern-0.8em\TeX}}}

\newcommand{\agent}{\mathcal{A}}
\newcommand{\macps}{MAS\xspace}

\newcommand{\controller}{\mathcal{C}}
\newcommand{\myipara}[1]{\vspace{0.6em} \noindent {\em #1.}}
\newcommand{\mypara}[1]{\vspace{0.6em} \noindent {\bf #1.}}
\newcommand{\astates}{X}
\newcommand{\ainputs}{U}
\newcommand{\aoutputs}{Y}
\newcommand{\atransitions}{T}
\newcommand{\astate}{x}
\newcommand{\ainput}{u}
\newcommand{\aoutput}{y}

\newcommand{\cstates}{Q}
\newcommand{\cinputs}{\Sigma}
\newcommand{\ctransitions}{\Delta}
\newcommand{\coutputs}{\Gamma}
\newcommand{\aid}{\mathsf{id}}
\newcommand{\coutput}{\gamma}
\newcommand{\cinput}{\sigma}

\IEEEoverridecommandlockouts                              

\title{Trust-aware Control for Intelligent Transportation Systems}

\author{Mingxi Cheng, Junyao Zhang, Shahin Nazarian, Jyotirmoy Deshmukh, Paul Bogdan
\thanks{The authors are with the Department of Electrical and Computer Engineering, University of Southern California, Los Angeles, CA, 90089, USA
(corresponding e-mail: pbogdan@usc.edu).}}

\begin{document}

\maketitle
\thispagestyle{empty}

\begin{abstract}
Many intelligent transportation systems are multi-agent systems, i.e.,
both the traffic participants and the subsystems within the
transportation infrastructure can be modeled as interacting agents.
The use of AI-based methods to achieve coordination among the
different agents systems can provide greater safety over
transportation systems containing only human-operated vehicles, and
also improve the system efficiency in terms of traffic throughput,
sensing range, and enabling collaborative tasks.  However, increased
autonomy makes the transportation infrastructure vulnerable to
compromised vehicular agents or infrastructure. This paper proposes a
new framework by embedding the trust authority into transportation infrastructure to systematically quantify the trustworthiness of agents
using an epistemic logic known as {\em subjective logic}. In this
paper, we make the following novel contributions: (\textit{i}) We propose a framework for using the quantified
trustworthiness of agents to enable trust-aware coordination and
control.  (\textit{ii}) We demonstrate how to synthesize trust-aware
controllers using an approach based on reinforcement learning.
(\textit{iii}) We comprehensively analyze an autonomous intersection
management (AIM) case study and develop a trust-aware version called
AIM-Trust that leads to lower accident rates in scenarios consisting
of a mixture of trusted and untrusted agents. 
\end{abstract}

\section{Introduction}\label{sec:intro}

Intelligent transportation systems are effectively multi-agent systems
(MAS), where both participants in the traffic as well as components of
the transportation infrastructure can be modeled as agents
that interact with each other \cite{shirazi2016looking}. For example, autonomous intersection
management \cite{dresner2004multiagent} consists of autonomous or
semi-autonomous vehicles that interact with an intersection manager,
while various systems such as adaptive platoons
\cite{milanes2013cooperative}, cooperative highway merging
\cite{rios2016automated, ntousakis2016optimal}, and cooperative
collision avoidance \cite{alonso2018cooperative,bin2017research} have
interacting vehicles that can be modeled as MAS. In the basic versions
of all such systems, the central focus is on the control algorithms
required to achieve the desired coordination objective. However,
increased level of autonomy renders such systems vulnerable to agents
whose functional behavior does not respect the assumptions made by the
coordination protocols. Agents can become compromised because they
could be the subjects of a malicious attack or simply because they
have defective sensors, actuators or control software, and thus threatens the entire system. The central
question we investigate in this paper is: {\em How can we guarantee
safety and performance of a multi-agent transportation system, when
participating agents are compromised?} 

While there has been significant emphasis on reasoning about the
security and privacy of MAS applications
\cite{sun2016cyber,alguliyev2018cyber,kamran2020risk}, using ideas from control
theory and cyber-security. These approaches tackle important problems
such as secure state estimation, attack detection and mitigation, and
system resilience. The view of most security-based approaches is
binary: either an agent is safe or compromised, and the mitigation
strategies are also thus restricted.  We argue that a transportation
system must remain operational even in the presence of compromised
agents, and in order to do so, we need a way to quantify the level of
{\em trustworthiness} of its constituent agents. Notions of
trustworthiness have been studied for vehicular \textit{ad hoc}
networks (e.g., \cite{hurl2019trupercept,hu2016replace,li2019trust}).
However, a formal definition and analysis of trustworthiness and how it can be systematically used to perform trust-aware control in MAS has received limited attention.
In previous work we proposed a general framework to tackle this problem but with limited details \cite{cheng2021general}. In this work, we provide trust evaluation and trust-aware control in intelligent transportation systems through a detailed case study of autonomous intersection control.



What makes an agent trustworthy? While this is a nuanced question, we
propose a mathematically precise definition of trustworthiness that
relies on two key principles: (\textit{i}) \textit{trusted} or
\textit{reliable agents} obey the control actions suggested by a
central (or distributed) coordinator, (\textit{ii}) \textit{trusted
agents} do not provide false information. 

We assume that a decision-making component in the transportation
infrastructure has an associated {\em trust authority} (TA) that can
evaluate an agent's trustworthiness through two kinds of observations:
direct observations and indirect observations gleaned from other
sources (e.g. other vehicles, other components of the infrastructure
such as road-side units). We call the latter local trust authorities
(LTAs). Each observation represents evidence that enhances the TA's
belief or disbelief in each agent (depending on whether the evidence
respectively indicates a desired or undesired behavior). For agents
that the TA has no opportunity to observe, the TA has no belief or
disbelief in the agent, but instead has uncertainty about its
trustworthiness. These ideas are rigorously developed in an {\em
epistemic} logic called {\em subjective logic} that we employ for
trust quantification and analysis.

\begin{figure}
    \centering
    \includegraphics[width=.9\columnwidth]{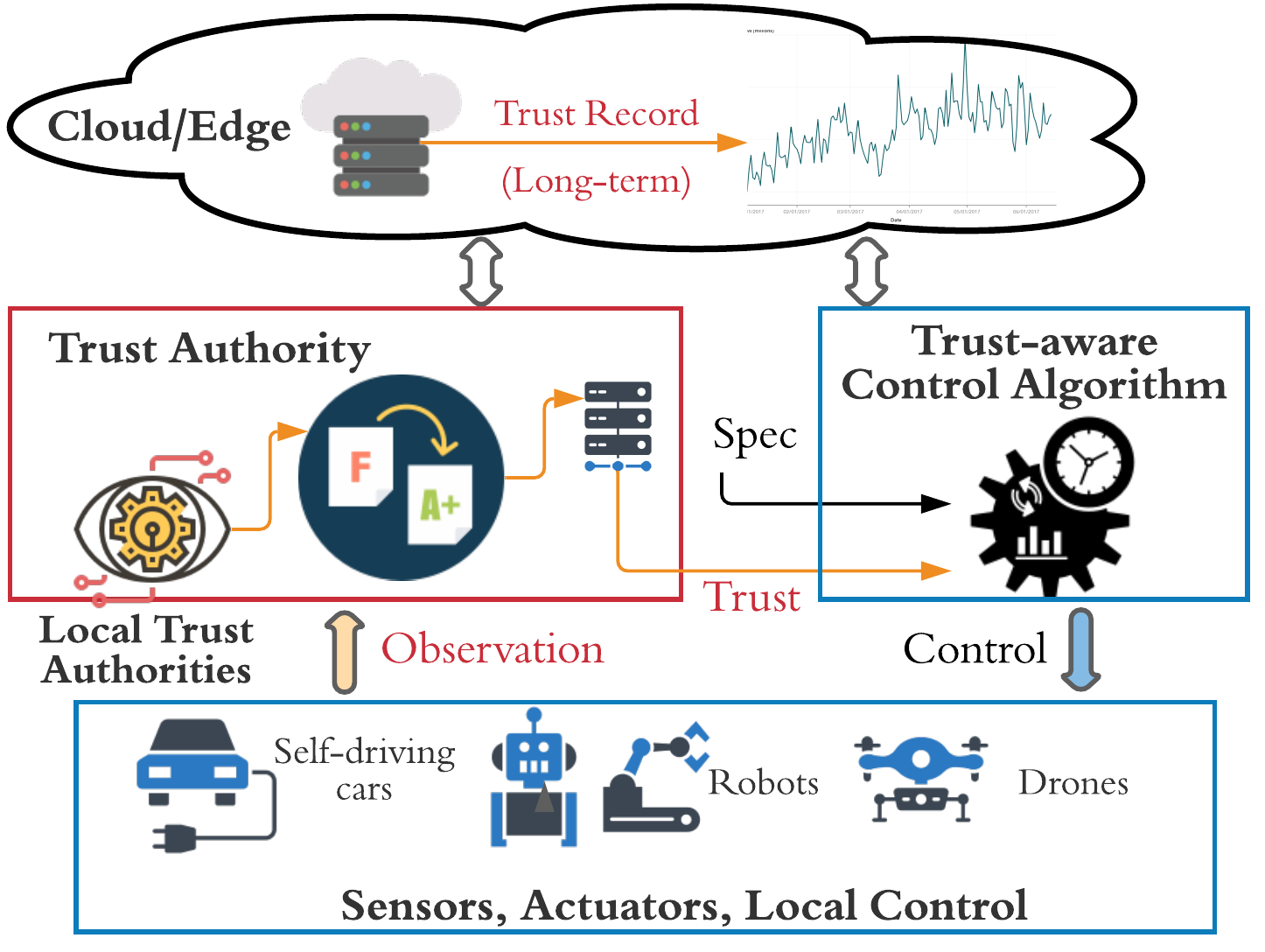}
    \caption{A trust-aware cyber-physical system.}
    \label{fig:CPS}
\end{figure}

We provide a conceptual depiction of a trust-aware MAS in
Fig.\ref{fig:CPS}. The system consists of a number of distributed
agents, each with their own sensors, actuators and local control
algorithms. The coordination among the agents is
achieved through either a centralized or a distributed control
algorithm. Traditionally, the only input to such a controller is the
global specification of desired behavior (encoding both mission
objectives for the MAS and safety constraints). In our framework, we
provide a trustworthiness score for each agent that is assumed to be
stored on a secure cloud/edge-based server. This score is computed by
a TA from observations of agents (reported through the agents' sensing
and communication modules). 
%
The trust-aware control algorithm uses the global specification describing desired
coordination behavior and trustworthiness scores to give control
actions, which the agents can choose to act on through their
decision-making, planning and actuation modules. 



In this paper, we perform a detailed analysis of our trust
quantification and trust-aware control framework on the autonomous
intersection management (AIM) system. AIM is a visionary system
proposed by Dresner and Stone \cite{dresner2008multiagent} to reduce
the traffic load of the current transportation infrastructure
and the resulting delays \cite{hausknecht2011autonomous}. It  
improves the intersection efficiency under the assumption 
that vehicles communicate with AIM and strictly follow the AIM
controller's instructions.  Compared to conventional intersection
control mechanisms such as stop signs and traffic signals, AIM 
policies provide consistently high traffic throughput while ensuring
the safety of the traffic participants \cite{hausknecht2011autonomous}.

In real-world scenarios, driving is fraught with uncertainties arising
from fallible as well as compromised human drivers, defective sensors
or actuators, noisy sensing environments, and
vehicle-to-infrastructure (V2X) communication. While existing AIM work
focuses on extending to scenarios with human-driven or semi-autonomous
vehicles \cite{sharon2017protocol,au2015autonomous}, it does not
account for the possibility of such systemic or adversarial
uncertainties. Thus, untrustworthiness arising out of uncertainty can
invalidate the benefits of AIM. A key contribution in this paper is to
showcase AIM as a detailed case study to highlight our algorithms for
trust quantification and trust-aware control for MAS. We show how we
can modify AIM to obtain AIM-Trust that uses trustworthiness scores of
traffic participants to generate trust-aware control actions. We
develop a reinforcement learning (RL) framework, where the action
space for the RL problem is the allocation of space-time buffers for
cars to navigate the intersection modulated by their trustworthiness,
and the reward space is defined in terms of collision freedom and
intersection throughput. Our formulation allows AIM-Trust to explore a
trade-off between performance (throughput) and safety (collision
avoidance). The main contributions are as follows:
\begin{itemize}[leftmargin=*]
    \item We propose a method to quantify trustworthiness scores for individual agents in a MAS.
    \item We provide a general and abstract framework for incorporating trust in generating control actions that guarantee the satisfaction
    of a global MAS specification.
    \item We showcase our trust quantification and trust-aware control
    methodology through a detailed case study of Autonomous Intersection 
    Management (AIM).
    \item We demonstrate how to embed the trustworthiness scores in a RL-based policy synthesis procedure. 
    \item Our empirical results show higher safety (at least $18.18\%$ and up to $89.28\%$ collision reduction) and efficiency ($15.53\%$ throughput improvement on average) of intersection management under the AIM-Trust controller compared to classical AIM that does not account for trustworthiness.
\end{itemize}




\section{Preliminaries}\label{sec:prelim}
\mypara{Multi-agent System Model}  A \macps is a
collection of dynamic agents interacting with each other and
with a controller (which could be centralized or distributed).  Each
agent $\agent$ can be described as a tuple
$(\aid,\astates,\ainputs,\aoutputs,\atransitions)$, where $\aid$ is a
unique positive integer, $\astates$ is a set of (internal) states of
the agent, $\ainputs$ is a set of inputs to the agent, $\aoutputs$ is
a set of observations provided by the agent, and $\atransitions$ is a
nondeterministic transition relation, subset of $\astates \times
\ainputs \times \astates \times \aoutputs$.  For every internal state
$\astate \in \astates$, the agent reads an input $\ainput \in
\ainputs$ and nondeterministically transitions to some state $\astate'
\in \astates$ providing output  $\aoutput \in \aoutputs$. We use the
set $\ainputs$ to model commands originated from the controller
(projected on this agent) and sensor inputs for the agent,
while the set $\aoutputs$ models the information shared by the agent
about its own state with other agents and the controller. We remark
that agents can also be modeled as having a stochastic transition
relation, where $\atransitions$ describes a probability
distribution of the next state and output conditioned on the current
state and input.

We assume that the set of agents interacting with the controller is dynamic and in any episode (defined as a finite period of time), agents can enter or leave the episode. 
An episode captures a time slice of the operation of the \macps.  The 
controller itself is also modeled as a tuple 
$(\cstates,\cinputs,\coutputs,\ctransitions)$, where $\cstates$ are the 
internal controller states, $\cinputs$ is the set of inputs read by the 
controller, $\coutputs$ are the command actions published by the controller
to all agents that are within the episode. The  {\em controller policy} 
$\ctransitions$ is a function from $\cstates\times\cinputs$ to 
$\cstates\times\coutputs$ that maps each controller state and input to its
next state and publishes output actions for the agents. 

\myipara{Global System Specification}
We assume that the global state space of the \macps is the product of the
state spaces of the controller and the agents involved in an episode. 
Essentially, at time $t$, the controller observes $\cinput(t)$ (which is a collection of $\aoutput(t)$ values of the available agents in the episode)
and publishes a (possibly empty)  control action $\coutput(t+1)$, and each 
agent processes the control action $\coutput(t)$ (projected to its $\aid$, i.e., $\ainput(t)$) and outputs $\aoutput(t+1)$. Both the controller and the 
agents also move to their respective next (internal) states upon executing 
internal actions. 
We assume that a global system specification is a property defined on the
space of the internal state trajectories of the agents. Such a property can
be easily specified in a multi-agent extension to any standard temporal logic
such as Signal Temporal Logic (STL) \cite{Maler2004}. 

\mypara{Background on Subjective Logic} To enable trustworthiness
evaluations, we utilize a probabilistic epistemic logic known as
\textit{subjective logic} (SL). SL is used in social systems to
quantify opinions and model trust relationships among humans.  In
general, SL is suitable for modeling and analyzing situations
involving uncertainty and relatively unreliable sources
\cite{josang2016subjective} and provides representations for opinions,
observations, and trust relationships. In our \macps context, each
agent $\agent$ is associated with an opinion and a corresponding trust
value, which are evaluated by the TA following specific rules. 

\begin{definition}[Opinion \cite{josang2016subjective}] \label{def:opinion}
Let $r$ be a quantity indicating the magnitude of {\em positive}
evidence obtained by the TA while observing the behavior an agent
$\agent$, and let $s$ be an analogous quantity indicating the
magnitude of {\em negative} evidence\footnote{For example, if the
agent violates (resp. satisfies) a behavioral specification, $s$
(resp. $r$) would quantify the degree of violation (resp.
satisfaction). For example if the global specifications is an STL
formula $\varphi$, $s$ (resp. $r$) is the (magnitude of) the robust
satisfaction value of $\varphi$ by the agent's behavior.} The binomial
opinion of an $\agent$ according to the TA is the set
$\overline{W}_{\agent}=\{b_{\agent},d_{\agent},u_{\agent},a_{\agent}\}$,
which consists of \textit{belief mass ($b_{\agent}$)},
\textit{disbelief mass ($d_{\agent}$)}, \textit{uncertainty mass
($u_{\agent}$)}, \textit{base rate ($a_{\agent}$)}, where $b_{\agent}
= \frac{r}{r+s+\omega}$, $d_{\agent} = \frac{s}{r+s+\omega}$,
$u_{\agent} = \frac{\omega}{r+s+\omega}$, $\omega = 2$ is a default
non-informative prior weight, satisfying the condition
$b_{\agent}+d_{\agent}+u_{\agent}=1$.
\end{definition}

Since we may have LTAs helping TA and they can observe agents and form opinions about a specific agent individually, we sometimes need to combine LTAs' opinions. Suppose our LTAs are trustworthy, then we can use \textit{cumulative fusion} operator to combine opinions as follows:
\begin{definition}[Cumulative Fusion \cite{josang2016subjective}] \label{def:fusion}
Suppose we have two LTAs $A$ and $B$, and they form opinions of agent $\agent$ as $\overline{W}^A_{\agent}= \{b^A_{\agent},d^A_{\agent},u^A_{\agent},a^A_{\agent}\}$ and $\overline{W}^B_{\agent}= \{b^B_{\agent},d^B_{\agent},u^B_{\agent},a^B_{\agent}\}$. The cumulative fusion, i.e., the combined opinion, $\overline{W}^{A\diamond B}_{\agent}$ of these two opinions is calculated as follows:
For $u^A_{\agent} \neq 0$ or $u^B_{\agent} \neq 0$:
\small
\begin{equation} \label{eq:cumulative_fusion}
    \begin{cases}
        b^{A\diamond B}_{\agent} =& (b^A_{\agent} u^B_{\agent}+b^B_{\agent} u^A_{\agent})/(u^A_{\agent}+u^B_{\agent}-u^A_{\agent} u^B_{\agent}),\\
        d^{A\diamond B}_{\agent} =& (d^A_{\agent} u^B_{\agent}+d^B_{\agent} u^A_{\agent})/(u^A_{\agent}+u^B_{\agent}-u^A_{\agent} u^B_{\agent}),\\
        u^{A\diamond B}_{\agent} =& (u^A_{\agent}u^B_{\agent})/(u^A_{\agent}+u^B_{\agent}-u^A_{\agent} u^B_{\agent}),\\
        a^{A\diamond B}_{\agent} =& (a^A_{\agent} u^B_{\agent}+a^B_{\agent} u^A_{\agent}-(a^A_{\agent}+a^B_{\agent})u^A_{\agent}u^B_{\agent})\\
        &/(u^A_{\agent}+u^B_{\agent}-2u^A_{\agent} u^B_{\agent}) \text{ if $u^A_{\agent}\neq1$ and $u^B_{\agent}\neq1$},\\
        a^{A\diamond B}_{\agent} =& (a^A_{\agent}+a^B_{\agent})/2 \text{ if $u^A_{\agent}=u^B_{\agent}=1$}.
    \end{cases}{}
\end{equation}{}
\end{definition}

\section{Trust Infrastructure}\label{sec:trustquant}
We consider a \macps consisting of a mixture of trustworthy and
untrustworthy agents and define a TA as an augmentation of the
controller $\controller$. Subject to an agent $\agent$, the TA observes
its behavior $\aoutputs$ and extracts knowledge (also known as
\textit{opinion} in SL) from observations (also known as
\textit{evidence}) and evaluates $\agent$'s trustworthiness as
follows:

\begin{definition}[Trustworthiness] \label{def:trust}
Given a TA and a specified agent $\agent$ in the \macps, the
\textit{trustworthiness} of $\agent$ assessed by TA is defined as
$p^{TA}_\agent = b^{TA}_\agent+u^{TA}_\agent*a^{TA}_\agent$
\cite{cheng2020there}, where  $b^{TA}_\agent$, $u^{TA}_\agent$,
and $a^{TA}_\agent$ are \textit{belief mass}, \textit{uncertainty
mass}, and \textit{base rate}, respectively.
\end{definition}

Besides the centralized trust authority TA, there might exist
distributed local trust authorities (LTAs) that help the TA to enlarge
observation range. Since in some cases, a single centralized trust
manager cannot cover all observation areas (e.g., in transportation
systems, imagine the TA as the department of motor vehicles and the TA
cannot directly observe all roads, hence, LTAs like road side units
serve as helpers and enlarge the observation range of TA). 

We envision a cloud-based (or edge-based) architecture as shown in
Fig. \ref{fig:CPS}, where centralized trust authority TA manages trust
histories of agents in an \macps in a hash table $\mathcal{H}$ on
cloud. TA regularly sends updates to and pull records from cloud. LTAs
report to TA but not directly talk to the cloud. In each trust
authority, an evidence measurement framework is embedded to evaluate
the behavior of agents to be positive $r$ or negative $s$ evidence.
Then the evidences are used to calculate the opinion as defined in
Definition \ref{def:opinion}.  Following our transportation system
example, road side units capture the undesired behaviors ($s$) of
vehicle $\agent$ and report to TA; the TA then updates the opinion of
$\agent$ using cumulative fusion operator defined in Definition
\ref{def:fusion} and push this record to the cloud and update
$\mathcal{H}$ to decrease the trustworthiness score of $\agent$.  In
the end, the trustworthiness measurement comes down to control
algorithms in $\controller$s as shown in Fig. \ref{fig:CPS}. The
control algorithm that takes trust as input is trust-aware, and makes
decisions with considerations of agents' historical behaviors.

\mypara{Trust-aware Control in \macps}
In the context of MAS, such as adaptive cruise control system
\cite{gong2019cooperative}, multi-agent autonomous traffic management
\cite{au2011enforcing}, and air/drone traffic control system
\cite{butler2017drone} the safety and behavior of one or a subset of
agents affects the efficiency and safety of the whole system. Such
systems are usually vulnerable to attacks that insert malicious agents
into the system for various purposes.  In these cases, a subjective
measurement is a must to identify malicious and untrustworthy agents.
Therefore, we propose to augment the controller's input space
$\cinputs$ to include the trustworthiness ($p$) of agents calculated
by the TA and enhance the controller policy $\ctransitions$ to be
trust-aware. 

\begin{figure}[!th]
    \centering
    \includegraphics[width=\columnwidth]{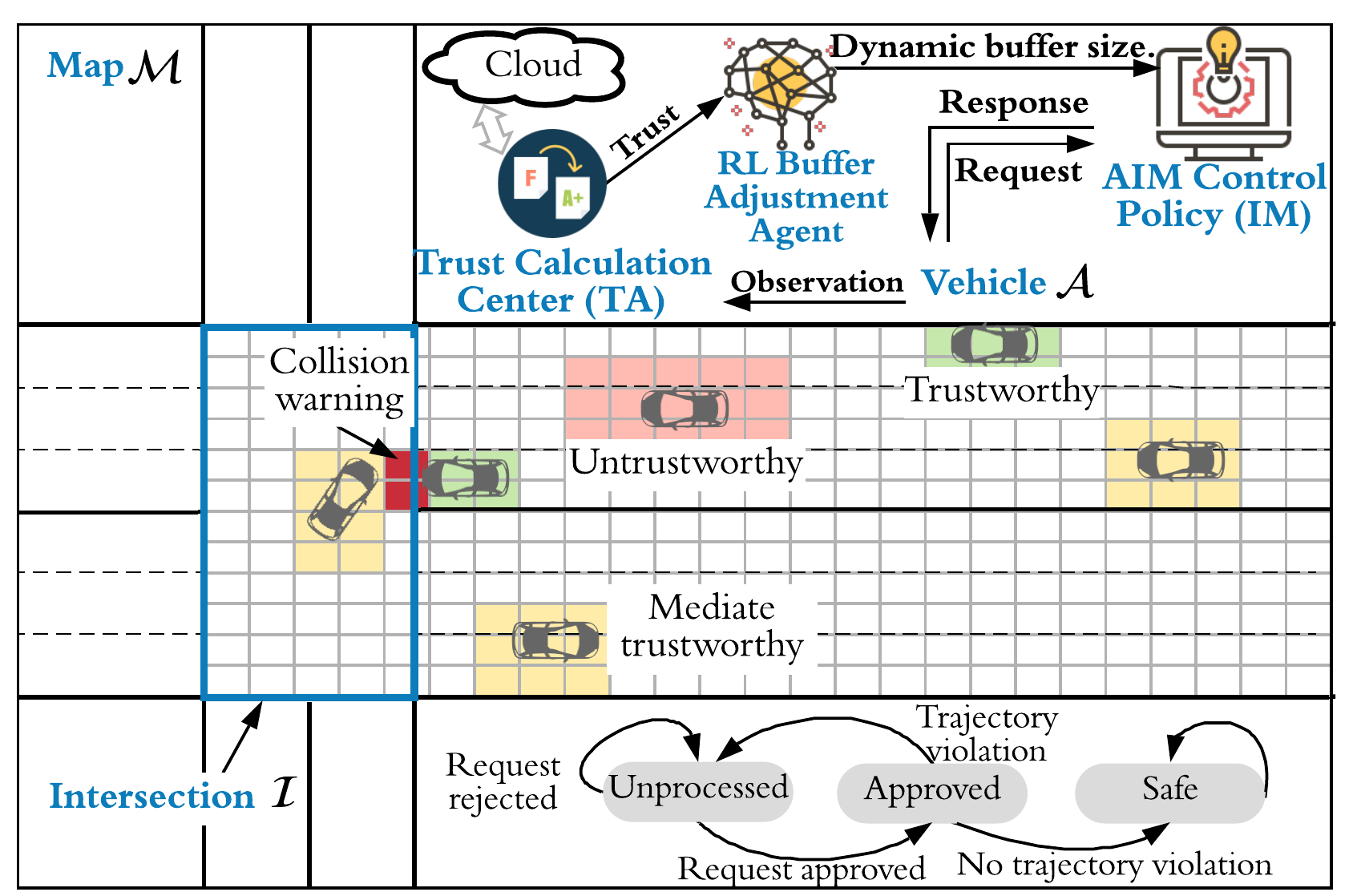}
    \captionsetup{font=small}
    \caption{A four-way intersection. Color-shaded areas represent the
    space-time buffers for each vehicle. Trustworthy vehicles have
    tight buffer since they are expected to obey the instructions with
    small error. Untrustworthy vehicles have large buffer because it
    is highly likely that they will act differently than instructed.
    Dark red area represents a collision warning in simulated
    trajectories. In this case, the vehicles are not permitted to
    enter the intersection and their requests are rejected.  The
    AIM-Trust framework consisting of IM and TA is shown on top.
    Detailed description of each component can be found in Algorithm
    \ref{alg:AIM-Trust}.}
\label{fig:AIM}
\end{figure}

\section{Case study in Autonomous Intersection Management}\label{sec:AIM}

\mypara{Autonomous Intersection Management (AIM)} The intersection traffic
in AIM is a simplified version of real-world intersection traffic.
Fig. \ref{fig:AIM} illustrates a four-way intersection example with
three lanes in each road leading to an intersection area $\mathcal{I}$
(marked by white doted rectangle).  A vehicle agent $\agent \in
\mathcal{V}$ on the road traveling towards but not already in the
intersection $\mathcal{I}$ is considered to be on the AIM \textit{map}
$\mathcal{M}$.  Any $\agent$ in $\mathcal{M}$ communicates with the
intersection manager (IM) $\controller$ by sending a request
$\aoutput(t)$ which consists of the vehicle identification number,
vehicle size, predicted arrival time, velocity, acceleration, arrival
and destination lanes, and then receives an instruction $\ainput(t)$. The
IM $\controller$ calculates the trajectory of $\agent$, makes a
grant or reject decision $\coutput(t)$ and sends the decision to
$\agent$. $\controller$ rejects a request if there exist conflicts in
the simulated trajectories. If $\controller$ approves the request,
{\em $\agent$ is responsible for obeying the instructions} to enter
and drive through $\mathcal{I}$. If $\controller$ rejects the request,
$\agent$ has to resend the request and await further instructions.

Important assumptions in AIM are as follows: (\textit{i}) For all
$\agent \in \mathcal{V}$, they follow the instructions of
$\controller$ strictly within an error tolerance. This restriction
guarantees safety by simulating trajectories and rejecting conflicting
requests.  (\textit{ii}) Each $\agent \in \mathcal{V}$ is associated
with a static {\em buffer size} with a minimum of value $1$. The
buffer size indicates the time-space reservation of $\agent$.
Trajectories are defined as conflicting if buffers of at least two vehicles
overlap (marked as dark red in Fig. \ref{fig:AIM} where shaded area
represents the buffer of each vehicle). The larger the buffer size,
the higher the safety, and the lower the efficiency or throughput.
Note that in conventional
AIM~\cite{dresner2008multiagent,sharon2017protocol}, all vehicles'
buffer sizes are set to $1$ and this preserves collision-freedom
because of assumption (\textit{i}).  However, assumption (\textit{i}) can be invalid as compromised agents can act
recklessly and disobey instructions, which can lead to collisions in
$\mathcal{I}$.  The small static buffer size in assumption
(\textit{ii}) intensifies this situation.

\mypara{AIM-Trust} \label{sec:AIM-Trust} Since in the real world, many
vehicles can be malicious and violate AIM assumptions, we associate
every IM with a TA based on our SL-based trust evaluation framework.
The TA uses the trustworthiness table $\mathcal{H}$ to obtain the
trustworthiness for each vehicle in $\mathcal{M}$. It uses these
values to make better trajectory approval decisions.  This framework
is called AIM-Trust.  We assume that TA-augmented IMs
(TA-IMs)\footnote{Note that in AIM-Trust, we denote a TA-augmented IM
as $\controller$ for simplicity, while it is in fact the combination
of a TA and an IM.} communicate with each other and they all
persistently maintain $\mathcal{H}$ through appropriate
synchronization mechanisms to maintain data coherence.  Each TA-IM
maintains a (coherent) local copy with part of $\mathcal{H}$ for
efficiency and scalability. In addition, road side units (RSUs) serve
as LTAs and provide coverage in places between the intersections to
track trustworthiness of vehicles.  AIM-Trust has two collision
avoidance mechanisms: (\textit{i}) a vehicle surveillance system to
identify untrustworthy behavior of vehicles (within $\mathcal{M}$),
and (\textit{ii}) an intelligent trust-based buffer adjusting
mechanism to help decrease collision risk while maintaining high
throughput. Fig. \ref{fig:AIM} and Algorithm \ref{alg:AIM-Trust} show
the operation details. Note that if there is no untrustworthy vehicle
in $\mathcal{M}$, AIM-Trust will reduce to original AIM algorithm with
fixed buffer size $1$ to ensure efficiency.

The TA-IM in AIM-Trust discriminates incoming vehicles into three
bins: \textit{unprocessed}, \textit{approved}, and \textit{safe} (see
Fig.~\ref{fig:AIM}).  A vehicle with its request unapproved by TA-IM
is \textit{unprocessed}. Once its request is approved, the vehicle is
\textit{approved} and under the surveillance of TA-IM for a few time
steps. If the vehicle behaves well, then it becomes \textit{safe} and
the surveillance ends.  However, if the vehicle violates the approved
trajectory with an intolerable error, then the vehicle goes back to
unprocessed bin and TA-IM starts processing its requests all over
again. Detailed state transition process can be found in Algorithm
\ref{alg:AIM-Trust}.  Compared to classical AIM where IM stops
interacting with a vehicle once the request is approved, AIM-Trust
includes a trust-based \textit{approve-observe} process to decide
whether to revoke and remake the approval decision. 
\SetAlCapNameFnt{\small}
\begin{algorithm}
\footnotesize
\captionsetup{font=small}
\caption{AIM-Trust algorithm. Vehicle $\agent$ sends a request to TA-IM $\controller$, which responds based on simulated trajectories.}\label{alg:AIM-Trust}
\SetKwInOut{Input}{Input}
\SetKwInOut{Output}{Output}
\SetKwBlock{kwPreproc}{Pre-process}{end}
\SetKwBlock{kwState}{State Transition}{end}
\Input{Vehicle $\agent$'s request message $\aoutput(t)$, i.e., vehicle identification number $\aid^\agent$, vehicle size, predicted arrival time, velocity, acceleration, arrival and destination lanes ($e^\agent, o^\agent$). }
\Output{Approve or reject decision of $\aoutput(t)$.}
\BlankLine
\kwPreproc { 
    Pre-process $\aoutput(t)$ for new reservation. \Comment{\textit{Same as AIM}}\\
    Trustworthiness $p^\agent \leftarrow trust\_calculator(\aid^\agent)$ \\
    Vehicle status $\xi^\agent \leftarrow unprocessed$ \\
} 
    
\kwState { 
    Buffer size $a^\agent \leftarrow buffer\_calculator(\aid^\agent,e^\agent,o^\agent,p^\agent)$\\
    Decision $\leftarrow AIM\_control\_policy(a^\agent)$
    \Comment{\textit{Same as AIM}}\\
    \textbf{Post-process }Send the decision to $\agent$. \Comment{\textit{Same as AIM}}\\
    \If {Decision == Approve}{$\xi^\agent \leftarrow approved$, run $surveillance(\aid^\agent)$ \\
    \lIf {$surveillance(\aid^\agent) == malicious$}{
    Go to \textbf{Pre-process}}
    $\xi^\agent \leftarrow safe$, $p^\agent \leftarrow trust\_calculator(\aid^\agent)$ once $\agent$ exits $\mathcal{M}$.\\
    }{}
} 
\end{algorithm}


\mypara{Trust Calculation} In AIM-Trust, TA-IM maintains a
trustworthiness / opinion table $\mathcal{H}$ and updates
$\mathcal{H}$ by either communicating with RSUs or considering new
evidences via cumulative fusion operator ($\diamond$) as shown in
Algorithm line $3$ and $13$.  The detailed procedure is shown in
Algorithm \ref{alg:Trust}.  The first trustworthiness update happens
when $\agent$ enters $\mathcal{M}$ and sends requests to manager
$\controller$:

\small
\begin{equation} \label{eq:trust_I}
\overline{W}^\controller_\agent = 
\begin{cases}
    \overline{W}^{UN}_\agent, \text{if vehicle $\agent$ is unknown,}\\
    \overline{W}^\controller_\agent, \text{if TA-IM(s) $\controller$ knows vehicle $\agent$,}\\
    \overline{W}^{LTA}_\agent,\text{if road side unit knows vehicle $\agent$},\\
    \overline{W}^{LAT\diamond \controller}_\agent,\text{if both LTA(s) and $\controller$ know vehicle $\agent$}.\\
\end{cases}
\end{equation}{}
\normalsize

\textbf{Case I.} ``$\agent$ is \textit{unknown}'' represents that the
vehicle does not have a record in $\mathcal{H}$.\footnote{We assume
$\overline{W}^{UN}_\agent=\{1,0,0,0.5\}$ to represent the maximum
belief based on autoepistemic logic~\cite{moore1985semantical} (the
vehicle is not reported to be untrustworthy, so it is trustworthy).} 

\textbf{Case II.}  ``$\controller$ knows $\agent$'' represents that
$\mathcal{H}$ has an entry of $\agent$ and $\agent$ has not been
picked up by RSUs after previous record update.

\textbf{Case III.}  Vehicle $\agent$ enters $\mathcal{M}$ and
$\controller$ receives LTA's report about $\agent$. Since RSUs are
only activated when undesired behavior happens, we expect
$\overline{W}^{LTA}_\agent$ to be a negative opinion.

\textbf{Case IV.}  RSU reports about $\agent$, which already has a
record in $\mathcal{H}$, hence, we use a cumulative fusion operator
($\diamond$) to merge these two opinions together.

\begin{algorithm}
\footnotesize
\caption{$trust\_calculator(\aid^\agent)$}\label{alg:Trust}
\uIf (\Comment{$\agent$ enters $\mathcal{M}$}){$\xi^\agent == unprocessed$}{$\overline{W}_\agent^{\controller}\leftarrow$ Eq. \ref{eq:trust_I} }
\uElseIf (\Comment{$\aoutput(t)$ approved and $surveillance(\aid^\agent) = malicious$}){$\xi^\agent == approved$}{
    $\overline{W}_\agent^{E}\leftarrow$ Def. \ref{def:opinion} \Comment{Evidence is collected before $\mathcal{I}$}\\
    $\overline{W}^{\controller}_\agent\leftarrow \overline{W}^{\controller\diamond E}_\agent$ 
    \\}
\uElseIf (\Comment{$\aoutput(t)$ approved and $surveillance(\aid^\agent) \neq malicious$} ){$\xi^\agent == safe$}{
    $\overline{W}_\agent^{E}\leftarrow$ Def. \ref{def:opinion} 
    \Comment{Evidence is collected in $\mathcal{I}$}\\
    $\overline{W}^{\controller}_\agent\leftarrow \overline{W}^{\controller\diamond E}_\agent$
    \Comment{$\agent$ exits $\mathcal{M}$}
    \\}{}
\KwRet $p^{\agent}\leftarrow p^{\controller}_\agent =b^{\controller}_\agent+u^{\controller}_\agent*a^{\controller}_\agent $\Comment{Definition \ref{def:trust}}
\end{algorithm}

This first updates of $\overline{W}^{\controller}_\agent $ and
$p^\agent$ are now completed and then used by buffer adjustment agent
as shown in Algorithm \ref{alg:AIM-Trust} line $7$.  Then, the AIM
control policy $\ctransitions$ generates accept / reject instruction.
After $\agent$ receives the instruction, the evidence framework starts
monitoring $\agent$'s behavior before it enters $\mathcal{I}$.  If
negative evidence is observed, then $\agent$ goes back to pre-process
as indicated in Algorithm \ref{alg:AIM-Trust} line $12$ and
$\overline{W}^{\controller}_\agent$ is updated as shown in Algorithm
\ref{alg:Trust} line $3$-$5$.  Otherwise, $\agent$ becomes safe and
proceeds to $\mathcal{I}$.  Once $\agent$ enters $\mathcal{I}$, the
surveillance system again observes $\agent$'s behavior and the
collision situation.  Positive / negative evidence based on collision
and trajectories is then evaluated and an opinion from evidence collected in $\mathcal{I}$
is derived as $\overline{W}^{E}_\agent$ (which is evaluated by
$\controller$ but we denote the superscript as $E$ to distinguish from
long-term $\overline{W}^{\controller}_\agent$).  After $\agent$ exits
$\mathcal{M}$, the trustworthiness of $\agent$ is updated again as
shown in Algorithm \ref{alg:Trust} line $6$-$8$ and uploaded to
$\mathcal{H}$.

\mypara{Evidence Measurement Framework} Exploiting the STL formalism,
we define a set of rules to specify a driving behavior to be desired
or undesired, i.e., quantify positive ($s$) or negative ($r$) evidence
for trust estimation. Desired / undesired behavior contribute to
positive / negative evidence; hence, they contribute to increasing /
deceasing of trustworthiness of an agent. Before the target driver
approaches $\mathcal{M}$, RSUs that have observed the target
vehicle assess the (undesired) behavior and generate (negative)
evidence. When the target vehicle arrives at the intersection,
AIM-Trust uses a self-embedded behavior measurement system to quantify
the target's behavior based on trajectory and collision status.

\newcommand{\alw}{\mathbf{G}}

\mypara{Evidence Evaluation at Road Side Units}
\label{sec:STL_RoadSide} Suppose an RSU observes vehicle $\agent$'s
trajectory and velocity, and the desired behavior is defined by a set
of rules, e.g., driving within one lane with negligible deviation and
under the designated speed limit. Hence, we define these properties
formally as follows: Subject to $\agent$, suppose the true trajectory
observed by RSU is $\cinput_{tr}$, the requested (or approved)
trajectory is $\coutput_{tr}$, the reported trajectory is
$\aoutput_{tr}$, and the negligible deviation or error is
$\epsilon_{tr}$.  Similarly, the observed speed of the vehicle is
$\cinput_{sp}$, and the designated speed is $\coutput_{sp}$, the
reported speed is $\aoutput_{sp}$, and the negligible error is
$\epsilon_{sp}$.\footnote{Note that in a \macps where all agents are
honest, the agent reported $\aoutput$ is the same as controller
observed $\cinput$.} We quantify $r$ and $s$ as follows:
\begin{equation} \label{eq:stl_roadside}
    \begin{cases}
         r = r + \beta_1, \text{ if } (\cinput_{tr},t) \models \varphi \wedge (\cinput_{sp},t) \models \psi;\\
        s = s + \beta_2, \text{ otherwise.}\\
    \end{cases}
\end{equation}

\begin{eqnarray}
\scriptstyle
\varphi \equiv & \scriptstyle \alw_{[t_1,t_2]}(|\cinput_{tr}(t) -\aoutput_{tr}(t)|\leq \epsilon_{tr} \wedge |\cinput_{tr}(t) -\coutput_{tr}(t)|\leq \epsilon_{tr})\\
\scriptstyle \psi \equiv & \scriptstyle \alw_{[t_1,t_2]}(|\cinput_{sp}(t) -\aoutput_{sp}(t)|\leq \epsilon_{sp} \wedge |\cinput_{sp}(t) -\coutput_{sp}(t)|\leq \epsilon_{sp})
\end{eqnarray}
These equations indicate that the true (observed) trajectory / speed
of a vehicle should not deviate from (\textit{i}) the requested
trajectory / speed and (\textit{ii}) the reported trajectory / speed
by more than $\epsilon_{tr}$ / $\epsilon_{sp}$ in time interval
$[t+t_1, t+t_2]$, where $t_1$, $t_2$, $\epsilon_{tr}$,
$\epsilon_{sp}$, $\beta_1$, and $\beta_2$ are hyper parameters.
$\beta_1,\beta_2 \in \Z^{+}$ are positive integers correlated with
values of $|\cinput(t) -\aoutput(t)|$ and $|\cinput(t)-\coutput(t)|$,
which indicates that the more proper / improper the behavior is, the
bigger the reward / penalty is.\footnote{In SL, $\beta_1$ and
$\beta_2$ usually takes value of $1$.}

\mypara{Evidence Evaluation at TA-IM} When vehicles are under the
surveillance of AIM-Trust before entering $\mathcal{I}$ (Algorithm
\ref{alg:AIM-Trust} line $11$), the evidence measurements in Eq.
\ref{eq:stl_roadside} are used to quantify the positive and negative
evidence. If negative evidence is observed, it means the vehicle
violates the approved trajectory by an intolerable error. (For
AIM-Trust, when TA-IM approves $\agent$'s requested trajectory,
$\aoutput=\coutput$.) Once vehicles enter $\mathcal{I}$, a new set of
rules are used to take into account the collision status of vehicles
and collisions in $\mathcal{I}$: $r = r + \beta_1$, if the vehicle
follows the approved trajectory and no collision happens; otherwise $s
= s + \beta_2$. 


\mypara{RL-based Buffer Adjustment Agent} AIM-Trust operates under the
assumption that there may exist untrustworthy vehicles who would not
follow instructions from TA-IM. Under such scenarios, the agents can not execute potential evasive maneuver to avoid the collisions, thereby these malicious agents with fixed and small buffer size will threat others agent, even the whole system. Then, how do we
determine optimal buffer size for agents with different trust values?
In order to assess this, and to have a buffer allocation policy, we
use reinforcement learning (RL) to explore the unknown environment and
figure out the appropriate buffer sizes.  In this section, we define
the RL formulation (deep Q-learning \cite{mnih2015human}) including
definitions of states, actions, and rewards.  In deep Q-learning, the
neural network is approximating a Q-learning table, where each entry
in the table is updated by $q(s_t, a_t) \leftarrow q(s_t, a_t) +\alpha
\left[r_{t+1} + \gamma \max_a q(s_{t+1}, a) - q(s_t, a_t)\right]$
\cite{dixit1990optimization}, where $s_t$ is state, $a_t$ is action,
$r_{t+1}$ is reward, $\alpha$ is learning rate, and $\zeta$ is
discounting factor. 

\myipara{State-State Transition-Action Spaces} We model a four-way
intersection with three lanes in each direction as shown in Fig.
\ref{fig:AIM}. To simulate the real world scenarios, we explicitly
allow vehicles on each lane to either go straight, turn left or right.
We define our states as $\boldsymbol{s_t} =
(\aid_t^1,e_t^1,o_t^1,p_t^1,...,\aid_t^n,e_t^n,o_t^n,p_t^n)^T$, where
$(\aid_t^i,e_t^i,o_t^i,p_t^i)$ are the vehicle identification number,
starting point, requested destination, and trustworthiness of vehicle
$i\in[1,n]$ at time $t$.  In each training time step $t$, vehicles
pass through $\mathcal{I}$ and fully exit $\mathcal{M}$.  Within one
episode, there are in total $\tau$ steps, which represent that the $n$
vehicles pass through $\tau$ intersections. $\forall i,t,\
(e_t^i,o_t^i)$ are randomly generated by the simulator, while $p_t^i$
is continuously updated based on Algorithm \ref{alg:Trust}.  The state
transition equations for the environment are defined as: $\aid_{t+1}^i
\leftarrow \aid_{t}^i$; $p_{t+1}^i \leftarrow
trust\_calculator(\aid^i_t)$; and $e_{t+1}^i, o_{t+1}^i \leftarrow
Random(\aid^i_{t+1})$, where $Random(\cdot)$ is the random starting
point and destination generator in simulator.  The action is defined
as $\boldsymbol{a_t} = (a_t^1,a_t^2,...,a_t^n)^T$, where $a_t^i$ is
the buffer size of vehicle $i$ at time $t$.  Neural network makes
prediction by assessing vehicles' positions, trustworthiness, and
requests. 

\myipara{Reward Function} The goal of our RL agent is to operate the
intersection with lowest collision rate and high throughput. It is
known that throughput is sensitive to buffer size, i.e., large buffer
size harms throughput. We take safety as our primary consideration and
improve the throughput under the promise of safety. Therefore, the
reward function reads:
\small
\begin{equation} \label{eq:aim-trust-reward}
    r_t^i = 
    \begin{cases}
        1 + \lambda(b_{th}-a_t^i),\quad \text{if no collision,}\\
        -(\tau-1)*[1 + \lambda(b_{th}-a_t^i)], \quad \text{otherwise,}
    \end{cases}    
\end{equation}
\normalsize
\noindent where $b_{th}$ is a hyper parameter indicating a reasonable
upper bound buffer size. $\lambda$ is a hyper parameter balances the
collision and throughput. The vehicle is removed once it collides and
not blocking $\mathcal{I}$, and is put back in
$\mathcal{M}$ in the next step. A training episode contains $\tau$ steps, and an episode ends once the maximum $\tau$ step size is reached. The
formulation of $r_t^i$ indicates that we want the buffer size to be as
small as possible to increase the throughput while penalizing
collisions.


\section{Experimental Results for AIM}\label{sec:results}
\begin{figure*}
    \includegraphics[width=.9\textwidth]{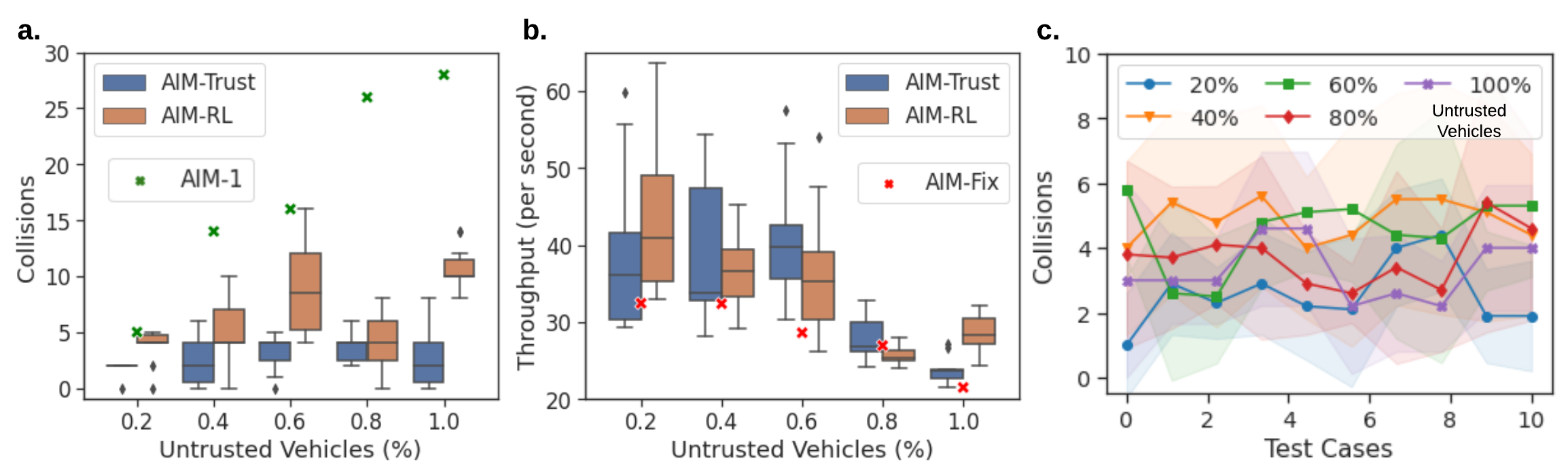}
    \captionsetup{font=small}
    \caption{\textbf{a.} Collision comparison between AIM-Trust, AIM-RL and AIM-1. 
    \textbf{b.} Throughput comparison between AIM-Trust, AIM-RL and
    AIM-Fix. Note in AIM-Trust and AIM-RL, the buffer size ranges from
    $0$ to $16$ in cases with $20\%$ to $60\%$ untrusted vehicles,
    while it ranges from $5$ to $21$ in cases with more untrusted
    vehicles (since the upper bound of $16$ is not enough for RL
    agents to learn a good collision avoidance strategy). This change
    of action space causes the discontinuity of trends in terms of
    collisions and throughput from $60\%$ and $80\%$ cases.
    \textbf{c.} Collision results of AIM-Trust with $10$ test cases
    that are different from the training set. Collision rates in test
    and training sets are consistent and stable even when $100\%$
    vehicles are untrustworthy.
    }
    \label{fig:AIM_10}
\end{figure*}

\begin{figure*}[!t]
    \centering
    \includegraphics[width=0.63\textwidth]{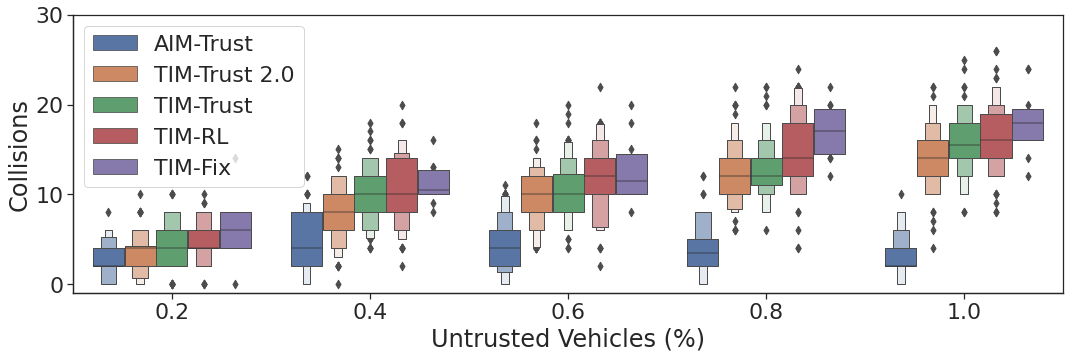}    \begin{adjustbox}{width=0.35\textwidth}
    \begin{tabular}[b]{cccccc}\\\toprule  
    UV & $20\%$ & $40\%$ & $60\%$ & $80\%$ & $100\%$ \\\midrule
    AIM-RL & $51.35\%$ & $50.00\%$ & $64.44\%$ & $18.18\%$ & $71.69\%$\\  
    AIM-1 & $64.00\%$ & $82.85\%$ & $79.37\%$ & $83.84\%$ & $89.28\%$\\
    TIM-RL & $44.46\%$ & $53.48\%$ & $60.94\%$ & $78.11\%$ & $79.85\%$\\
    TIM-Fix & $58.70\%$ & $56.51\%$ & $64.33\%$ & $78.11\%$ & $81.55\%$\\
    TIM-Trust & $42.98\%$ & $52.31\%$ & $57.22\%$ & $71.18\%$ & $78.48\%$ \\
    TIM-Trust $2.0$ & $25.14\%$ & $36.58\%$ &  $54.28\%$ & $70.61\%$ & $76.86\%$ \\\midrule
    AIM-Fix & $16.93\%$ & $17.25\%$ & $30.13\%$ & $3.61\%$ & $9.77\%$ \\ \bottomrule
    &&&&&\\
    &&&&&\\
    &&&&&\\
    \end{tabular}
    \end{adjustbox}
    \captionlistentry[table]{}
    \captionsetup{font=small}
    \caption{Collision comparison. 
    The results of RL-based methods contain $10$ test cases in each
    scenario (untrusted vehicle percentage varies from $20\%$ to
    $100\%$), and $10$ runs of each case (hence in total $100$ data
    points in each box). Trustworthiness-aware methods have lower
    collisions in all scenarios.  Table 1: AIM-Trust's collision rate
    decrements compared to baselines, and AIM-Trust's throughput
    increments compared to AIM-Fix. UV indicates the untrusted vehicle
    percentage.}
    \label{fig:TIM}
\end{figure*}
    
\mypara{Experiment Setup} We consider in an RL
training episode, $n=10$ vehicles pass through $\tau=10$ intersections
and we monitor the collisions occuring within these $n \tau =100$
passings as $c$.  For each intersection (or step) $t$ in an episode,
$n$ vehicles enter and leave the intersection following randomly
picked start points, $[e^1_t,...,e^n_t]$, and destinations,
$[o^1_t,...,o^n_t]$.  We generate $1$ set of starting points and
destinations as training set, and generate $10$ independent sets as a
test set.  We consider two performance metrics: collision rate
$\frac{c}{n\tau}$, and throughput $\frac{n\tau-c}{\mathcal{T}}$, where
$\mathcal{T}$ is the time (in seconds) elapsed in one episode.  All
simulations are done in the AIM simulator \cite{AIMsim}, and we
provide video demonstrations in \cite{demos}. 

\myipara{Baselines} \label{sec:exp_baseline} We first compare our
proposed AIM-Trust with $3$ AIM family baselines: the original AIM
algorithm with fixed buffer size $1$, namely AIM-1; the modified AIM
algorithm with fixed averaging buffer size, namely AIM-Fix (we
manually select the fixed buffer size for this baseline to force it
perform similarly as AIM-Trust in terms of collision rate, then
compare the throughput with AIM-Trust); and a variation of AIM-Trust
without considering the trust factor, $p$, in the state space, namely
AIM-RL. 

In addition, we compare AIM-Trust with traffic light-based
intersection control methods which are not in AIM family. We follow
\cite{liang2018deep} to construct a deep reinforcement learning
(DRL)-based traffic light cycle control method as $\controller$ to
operate the intersection. We denote this method as TIM-RL, i.e.,
traffic-light intersection management based on RL.  Since TIM methods
focus on operating the intersection with high efficiency without
considering untrustworthy agents, they have no collision avoidance
mechanism.  To make the baseline more competitive in the scenarios
involving untrustworthy vehicles, we propose two enhanced TIM methods:
(\textit{i}) TIM-Trust, which includes trustworthiness $p^i_t$ in
TIM-RL's state space, and (\textit{ii}) TIM-Trust $2.0$, which
includes trustworthiness and has collision penalties in reward
function.  Furthermore, we include a fixed cycle traffic light (no RL
agent involved) to replicate conventional traffic light control method
for comparison, which is denoted as TIM-Fix.

\mypara{Collision Results}
In this section, we first present collision comparison results in AIM
family as shown in Fig. \ref{fig:AIM_10}a.  We vary the percentage of
untrusted vehicles from $20\%$ to $100\%$.  Since RL training embeds
the randomness from initialization naturally, we train AIM-Trust $10$
times and report the mean-variance results to show the stability.
Fig. \ref{fig:AIM_10}a shows that AIM-Trust decreases the collision
numbers drastically compared to AIM-1 (see Table 1 for detailed
numerical results).  Since AIM cannot deal with the violation of
assumption (\textit{i}) (as described in Section \ref{sec:AIM}), the
small and fixed buffer size leads to high collision rate. The more
untrustworthy vehicles in the system, the more the collisions.
AIM-Trust's adjustable buffer size helps to decrease the collision
rate and maintains stable low collision rate even when all vehicles
are untrustworthy.

In order to examine the effectiveness of the trustworthiness, we make a baseline AIM-RL by taking out the trust factor from AIM-Trust. Except the trust factor, AIM-RL is exactly the same as AIM-Trust with adjustable buffer size to decrease the collision rate.
In addition, we control the training process of AIM-RL and AIM-Trust
to be the same to ensure fair comparison. The experimental results in
Fig.~\ref{fig:TIM} and Table 1 demonstrate that the trustworthiness of
a vehicle is key to infer the appropriate buffer size. To
investigate the convergence and robustness of the AIM-Trust agent, we
consider the collision performance of AIM-Trust in test set and
training set is consistent as shown in Fig. \ref{fig:AIM_10}c, which
indicates that pre-trained AIM-Trust performs well in unseen
traffic scenarios.

Next, we show the performance of AIM-Trust compared with non-AIM
methods, TIM-Fix, TIM-RL, TIM-Trust, and TIM-Trust $2.0$, in Figure
\ref{fig:TIM} and Table 1.  Compared with conventional traffic
light-based intersection control methods, AIM-Trust is advantageous
since it considers the uncertainty and trustworthiness of vehicles,
and decreases collision rate by foreseeing the potential trajectories
of trustworthy and untrustworthy vehicles.  To demonstrate the
significance and advantage of proposed trustworthiness of agents, we
enhance TIM-RL and reveal that trust-based methods, TIM-Trust and
TIM-Trust $2.0$, beat TIM-RL in terms of collision rate in all
scenarios.  Experimental results confirm that in a \macps, when there
exist untrustworthy agents, trustworthiness is important for control
algorithms to infer the involved uncertainty.

\mypara{Throughput Results} In addition to collision rate, we compare
the throughput between AIM-Trust and AIM-Fix.  For a fair comparison,
we let buffer size of AIM-Fix to be $[9,9.5,11,13,21.5]$ under $20\%$
to $100\%$ untrusted vehicles such that AIM-Fix has similar (slightly
larger) collision rates as AIM-Trust. Then, under similar collision
rate, we compare the throughput.  As shown in Fig. \ref{fig:AIM_10}b
and Table 1, AIM-Trust's throughput improvement demonstrates that the
RL-based buffer adjustment not only decreases the collision rate, but
also benefits the throughput. Compared to AIM-Fix, AIM-Trust on
average achieves higher throughput in all cases. Note that the
collision rate of AIM-Fix is higher than AIM-Trust and based on
throughput calculation, high collision rate actually gives advantages.
With different collision rate, the comparison of throughput is unfair
since the sacrifice of safety generates better performance in
throughput (note that in simulation we remove the collision vehicle
immediately from the map and does not effect the traffic flow).  To
compare with AIM-RL fairly, we can compare with $80\%$ untrusted
vehicle as AIM-Trust and AIM-RL achieve similar collision rate in this
case; and the throughput results show AIM-Trust achieves higher
throughput.  In other words, AIM-Trust with both lower collision rate
and higher throughput indicates that AIM-Trust is much better than
AIM-Fix and AIM-RL; and the trust factor we defined in this work is
the major contributor of this out-performance.

\mypara{Collision-free AIM-Trust} AIM-Trust with the throughput
consideration provides significant collision rate reduction compared
to AIM-1.  However, it cannot guarantee collision-free due to the
safety and throughput dilemma.  In AIM-Trust, we consider collision
and throughput via balancing factor $\lambda$ in the reward function
Eq. \ref{eq:aim-trust-reward}.  To demonstrate that AIM-Trust can
deduce appropriate buffer sizes based on trust, we relax
the throughput requirement and modify the reward
function of AIM-Trust to be $r_t^i=1$ if no collision and $r_t^i=40$
otherwise. We also let RL agent choose buffer size in range $0$ to
$26$ (we denote this new version as AIM-Trust $2.0$). Through these
minor modifications, AIM-Trust $2.0$ focuses on collision avoidance
and learns to achieve collision-free in training. On average, the resulting buffer sizes of AIM-Trust $2.0$ in cases with $20\%$ to $100\%$ untrustworthy vehicles are $[14.4, 14, 14.4, 20, 24]$. With same reward function, AIM-RL
$2.0$ (i.e., AIM-Trust $2.0$ without trust factor in state space) also
learns to avoid collision completely, but with higher buffer sizes
$[14, 15, 16, 20, 26]$ that lead to lower throughput. 

\begin{figure}[!t]
    \centering
    \includegraphics[width=\columnwidth]{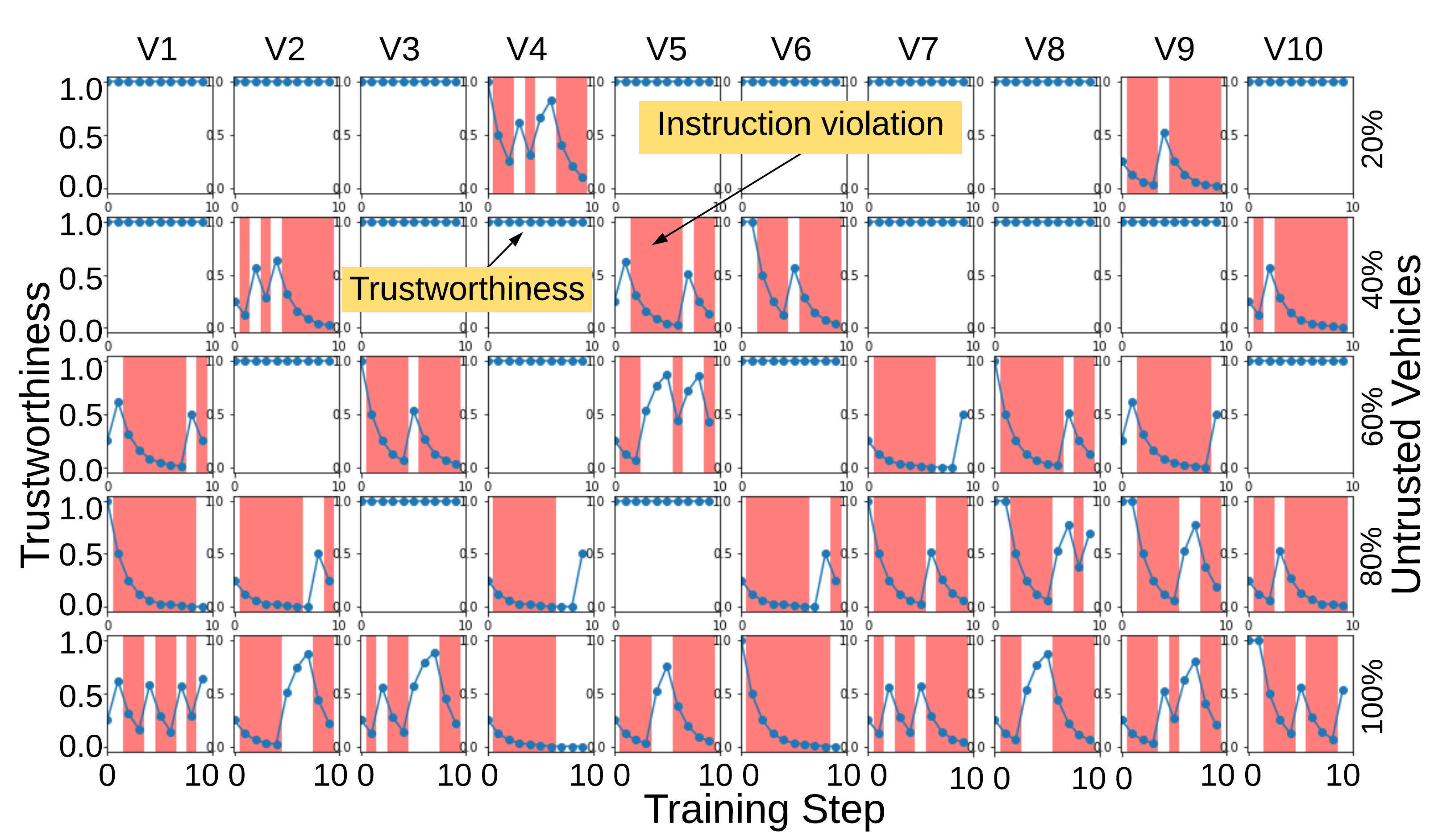}
    \captionsetup{font=small}
    \caption{Trustworthiness results (blue lines) and instruction
    violation results (red areas). In $20\%$ untrusted case, $2$ of
    $10$ vehicles may or may not follow instructions, hence, $2$
    figures in the first row contain red areas. Our trust calculation
    precisely captures the instruction violation: a vehicle's
    trustworthiness increases when it follows the instruction, and
    decreases otherwise.}
    \label{fig:trust}
\end{figure}

\mypara{Trust Results} Here, we present the trustworthiness
quantification of vehicles in our experiments. Fig. \ref{fig:trust} shows the trustworthiness results
of $10$ vehicles in one of our experiments. Each column corresponds to
a vehicle, which may or may not be trustworthy. Each row represents a
full episode of training with $10$ time steps (i.e., vehicles passing
through $10$ intersections). For example, the first row contains the
trustworthiness evaluations of all $10$ vehicles passing through $10$
intersections, and $20\%$ of them are untrustworthy. 
Red areas indicate that a vehicle does not follow the approved trajectory in simulation, and this causes
trustworthiness (blue line) decrements. These results show that our trustworthiness evaluation accurately captures the undesired behavior of vehicles, and is significantly helpful when used in control algorithms.

\section{Conclusion}\label{sec:conclusion}

AIM is designed to provide a collision-free intersection management
with high throughput.
However, in the application environment where
there exists untrustworthy even malicious vehicles that do not follow
the instructions, conventional AIM leads to a large amount of
collisions (up to $28\%$).  This example reveals the need for
trustworthiness measure in MAS we proposed in this work.  We design a
trust evaluation framework and propose to use evaluated
trustworthiness in control algorithms. We demonstrate in a case study
how to embed trustworthiness in intersection management by
designing AIM-Trust. 
To evaluate the effectiveness of the trust factor, we explicitly compare our AIM-Trust with baselines, and the experimental
results show that the trust factor reduces the collision rate in all
cases. In addition, in trustworthiness results, we directly see that
trust scores accurately reflect the behavior quality of vehicles. For
future work, we would like to refine our trust framework to be more
comprehensive and applicable to a broader range of MAS control
algorithms.

\section{Acknowledgment}
The authors gratefully acknowledge the support by the National Science Foundation under the Career Award CPS/CNS-1453860, the Career Award SHF-2048094, the NSF awards under Grant Numbers CCF-1837131, MCB-1936775, CNS-1932620, SHF-1910088, CPS/CNS-2039087, CMMI 1936624 and the DARPA Young Faculty Award and DARPA Director's Fellowship Award, under Grant Number N66001-17-1-4044, a Northrop Grumman grant, and a grant from Toyota R\&D North America. The views, opinions, and/or findings contained in this article are those of the authors and should not be interpreted as representing the official views or policies, either expressed or implied by the Defense Advanced Research Projects Agency, the Department of Defense or the National Science Foundation.

\bibliographystyle{IEEEtran}
\bibliography{sample-base.bib,reference.bib}

\begin{thebibliography}{10}
\providecommand{\url}[1]{#1}
\csname url@samestyle\endcsname
\providecommand{\newblock}{\relax}
\providecommand{\bibinfo}[2]{#2}
\providecommand{\BIBentrySTDinterwordspacing}{\spaceskip=0pt\relax}
\providecommand{\BIBentryALTinterwordstretchfactor}{4}
\providecommand{\BIBentryALTinterwordspacing}{\spaceskip=\fontdimen2\font plus
\BIBentryALTinterwordstretchfactor\fontdimen3\font minus
  \fontdimen4\font\relax}
\providecommand{\BIBforeignlanguage}[2]{{%
\expandafter\ifx\csname l@#1\endcsname\relax
\typeout{** WARNING: IEEEtran.bst: No hyphenation pattern has been}%
\typeout{** loaded for the language `#1'. Using the pattern for}%
\typeout{** the default language instead.}%
\else
\language=\csname l@#1\endcsname
\fi
#2}}
\providecommand{\BIBdecl}{\relax}
\BIBdecl

\bibitem{shirazi2016looking}
M.~S. Shirazi and B.~T. Morris, ``Looking at intersections: a survey of
  intersection monitoring, behavior and safety analysis of recent studies,''
  \emph{IEEE Transactions on Intelligent Transportation Systems}, vol.~18,
  no.~1, pp. 4--24, 2016.

\bibitem{dresner2004multiagent}
K.~Dresner and P.~Stone, ``Multiagent traffic management: A reservation-based
  intersection control mechanism,'' in \emph{Proceedings of the Third
  International Joint Conference on Autonomous Agents and Multiagent
  Systems-Volume 2}, 2004, pp. 530--537.

\bibitem{milanes2013cooperative}
V.~Milan{\'e}s, S.~E. Shladover, J.~Spring, C.~Nowakowski, H.~Kawazoe, and
  M.~Nakamura, ``Cooperative adaptive cruise control in real traffic
  situations,'' \emph{IEEE Transactions on intelligent transportation systems},
  vol.~15, no.~1, pp. 296--305, 2013.

\bibitem{rios2016automated}
J.~Rios-Torres and A.~A. Malikopoulos, ``Automated and cooperative vehicle
  merging at highway on-ramps,'' \emph{IEEE Transactions on Intelligent
  Transportation Systems}, vol.~18, no.~4, pp. 780--789, 2016.

\bibitem{ntousakis2016optimal}
I.~A. Ntousakis, I.~K. Nikolos, and M.~Papageorgiou, ``Optimal vehicle
  trajectory planning in the context of cooperative merging on highways,''
  \emph{Transportation research part C: emerging technologies}, vol.~71, pp.
  464--488, 2016.

\bibitem{alonso2018cooperative}
J.~Alonso-Mora, P.~Beardsley, and R.~Siegwart, ``Cooperative collision
  avoidance for nonholonomic robots,'' \emph{IEEE Transactions on Robotics},
  vol.~34, no.~2, pp. 404--420, 2018.

\bibitem{bin2017research}
F.~Bin, F.~XiaoFeng, and X.~Shuo, ``Research on cooperative collision avoidance
  problem of multiple uav based on reinforcement learning,'' in \emph{2017 10th
  International Conference on Intelligent Computation Technology and Automation
  (ICICTA)}.\hskip 1em plus 0.5em minus 0.4em\relax IEEE, 2017, pp. 103--109.

\bibitem{sun2016cyber}
C.-C. Sun, C.-C. Liu, and J.~Xie, ``Cyber-physical system security of a power
  grid: State-of-the-art,'' \emph{Electronics}, vol.~5, no.~3, p.~40, 2016.

\bibitem{alguliyev2018cyber}
R.~Alguliyev, Y.~Imamverdiyev, and L.~Sukhostat, ``Cyber-physical systems and
  their security issues,'' \emph{Computers in Industry}, vol. 100, pp.
  212--223, 2018.

\bibitem{kamran2020risk}
D.~Kamran, C.~F. Lopez, M.~Lauer, and C.~Stiller, ``Risk-aware high-level
  decisions for automated driving at occluded intersections with reinforcement
  learning,'' in \emph{2020 IEEE Intelligent Vehicles Symposium (IV)}.\hskip
  1em plus 0.5em minus 0.4em\relax IEEE, 2020, pp. 1205--1212.

\bibitem{hurl2019trupercept}
B.~Hurl, R.~Cohen, K.~Czarnecki, and S.~Waslander, ``Trupercept: Trust
  modelling for autonomous vehicle cooperative perception from synthetic
  data,'' in \emph{2020 IEEE Intelligent Vehicles Symposium (IV)}.\hskip 1em
  plus 0.5em minus 0.4em\relax IEEE, pp. 341--347.

\bibitem{hu2016replace}
H.~Hu, R.~Lu, Z.~Zhang, and J.~Shao, ``Replace: A reliable trust-based platoon
  service recommendation scheme in vanet,'' \emph{IEEE Transactions on
  Vehicular Technology}, vol.~66, no.~2, pp. 1786--1797, 2016.

\bibitem{li2019trust}
F.~Li, D.~Mikulski, J.~R. Wagner, and Y.~Wang, ``Trust-based control and
  scheduling for ugv platoon under cyber attacks,'' SAE Technical Paper, Tech.
  Rep., 2019.

\bibitem{cheng2021general}
M.~Cheng, C.~Yin, J.~Zhang, S.~Nazarian, J.~Deshmukh, and P.~Bogdan, ``A
  general trust framework for multi-agent systems,'' in \emph{Proceedings of
  the 20th International Conference on Autonomous Agents and MultiAgent
  Systems}, 2021, pp. 332--340.

\bibitem{dresner2008multiagent}
K.~Dresner and P.~Stone, ``A multiagent approach to autonomous intersection
  management,'' \emph{Journal of artificial intelligence research}, vol.~31,
  pp. 591--656, 2008.

\bibitem{hausknecht2011autonomous}
M.~Hausknecht, T.-C. Au, and P.~Stone, ``Autonomous intersection management:
  Multi-intersection optimization,'' in \emph{2011 IEEE/RSJ International
  Conference on Intelligent Robots and Systems}.\hskip 1em plus 0.5em minus
  0.4em\relax IEEE, 2011, pp. 4581--4586.

\bibitem{sharon2017protocol}
G.~Sharon and P.~Stone, ``A protocol for mixed autonomous and human-operated
  vehicles at intersections,'' in \emph{International Conference on Autonomous
  Agents and Multiagent Systems}.\hskip 1em plus 0.5em minus 0.4em\relax
  Springer, 2017, pp. 151--167.

\bibitem{au2015autonomous}
T.-C. Au, S.~Zhang, and P.~Stone, ``Autonomous intersection management for
  semi-autonomous vehicles,'' in \emph{Routledge Handbook of
  Transportation}.\hskip 1em plus 0.5em minus 0.4em\relax Routledge, 2015, pp.
  116--132.

\bibitem{Maler2004}
O.~Maler and D.~Nickovic, ``Monitoring temporal properties of continuous
  signals,'' in \emph{FORMATS/FTRTFT}, 2004.

\bibitem{josang2016subjective}
A.~J{\o}sang, \emph{Subjective logic}.\hskip 1em plus 0.5em minus 0.4em\relax
  Springer, 2016.

\bibitem{cheng2020there}
M.~Cheng, S.~Nazarian, and P.~Bogdan, ``There is hope after all: Quantifying
  opinion and trustworthiness in neural networks,'' \emph{Frontiers in
  Artificial Intelligence}, vol.~3, p.~54, 2020.

\bibitem{gong2019cooperative}
S.~Gong, A.~Zhou, and S.~Peeta, ``Cooperative adaptive cruise control for a
  platoon of connected and autonomous vehicles considering dynamic information
  flow topology,'' \emph{Transportation Research Record}, vol. 2673, no.~10,
  pp. 185--198, 2019.

\bibitem{au2011enforcing}
T.-C. Au, N.~Shahidi, and P.~Stone, ``Enforcing liveness in autonomous traffic
  management,'' in \emph{Twenty-Fifth AAAI Conference on Artificial
  Intelligence}, 2011.

\bibitem{butler2017drone}
E.~K. Butler, A.~A. Chandra, P.~R. Chowdhary, S.~M. Glissmann-Hochstein, T.~D.
  Griffin, D.~Jadav, S.~Lee, and H.~R. Strong~Jr, ``Drone air traffic control
  and flight plan management,'' Dec.~26 2017, uS Patent 9,852,642.

\bibitem{moore1985semantical}
R.~C. Moore, ``Semantical considerations on nonmonotonic logic,''
  \emph{Artificial intelligence}, vol.~25, no.~1, pp. 75--94, 1985.

\bibitem{mnih2015human}
V.~Mnih, K.~Kavukcuoglu, D.~Silver, A.~A. Rusu, J.~Veness, M.~G. Bellemare,
  A.~Graves, M.~Riedmiller, A.~K. Fidjeland, G.~Ostrovski \emph{et~al.},
  ``Human-level control through deep reinforcement learning,'' \emph{nature},
  vol. 518, no. 7540, pp. 529--533, 2015.

\bibitem{dixit1990optimization}
A.~K. Dixit, J.~J. Sherrerd \emph{et~al.}, \emph{Optimization in economic
  theory}.\hskip 1em plus 0.5em minus 0.4em\relax Oxford University Press on
  Demand, 1990.

\bibitem{AIMsim}
``Aim4 1.0-snapshot api,''
  \url{http://www.cs.utexas.edu/~aim/aim4sim/aim4-release-1.0.3/aim4-root/target/site/apidocs/index.html},
  accessed: 2020-07-26.

\bibitem{demos}
\BIBentryALTinterwordspacing
``Video demonstrations for aim-trust.'' [Online]. Available:
  \url{https://drive.google.com/drive/folders/1xYcms9UKM0Z5yq8IBjzeI_G5GHd1kXXo?usp=sharing}
\BIBentrySTDinterwordspacing

\bibitem{liang2018deep}
X.~Liang, X.~Du, G.~Wang, and Z.~Han, ``Deep reinforcement learning for traffic
  light control in vehicular networks,'' \emph{arXiv preprint
  arXiv:1803.11115}, 2018.

\end{thebibliography}

\end{document}